\documentclass{article}
\usepackage[square,numbers]{natbib}
\usepackage{color}
\usepackage{bm}
\usepackage{arydshln}
\usepackage{subcaption}
\usepackage{graphicx}
\usepackage[final]{nips_2016}

\usepackage[utf8]{inputenc} % allow utf-8 input
\usepackage[T1]{fontenc}    % use 8-bit T1 fonts
\usepackage{hyperref}       % hyperlinks
\usepackage{url}            % simple URL typesetting
\usepackage{booktabs}       % professional-quality tables
\usepackage{amsfonts}       % blackboard math symbols
\usepackage{nicefrac}       % compact symbols for 1/2, etc.
\usepackage{microtype}      % microtypography

\title{On Multiplicative Integration with\\ Recurrent Neural Networks}

\author{\hspace{-9pt}Yuhuai Wu$^{1,*}$, Saizheng Zhang$^{2,}$\thanks{Equal contribution.}\ \ , Ying Zhang$^2$, Yoshua Bengio$^{2,4}$ and Ruslan Salakhutdinov$^{3,4}$\\
 \hspace{-9pt} $^1$University of Toronto, $^2$MILA, Universit\'e de Montr\'eal, $^3$Carnegie Mellon University, $^4$CIFAR\\
 \hspace{-9pt} \texttt{ywu@cs.toronto.edu,$^2$\{firstname.lastname\}@umontreal.ca,rsalakhu@cs.cmu.edu} \\
}

\begin{document}
% \nipsfinalcopy is no longer used

\maketitle

\begin{abstract}
We introduce a general and simple structural design called
``Multiplicative Integration''~(MI) to improve  
recurrent neural networks (RNNs).
MI changes the way in which information from difference sources flows
and is integrated in the
computational building block of an RNN,
while introducing almost no extra parameters.
The new structure can be easily embedded into many popular
RNN models, including LSTMs and GRUs.
We empirically analyze its learning behaviour and 
conduct evaluations on several tasks using different RNN models.
Our experimental results demonstrate that Multiplicative Integration
can provide a substantial performance boost %leads to better performances 
over many of the existing RNN models.
\end{abstract}
\section{Introduction}
\label{sec:intro}
%There has been a resurgence of Recurrent Neural Networks (RNNs) that 
%achieve promising results on many complex sequence 
%learning tasks
%, such as neural machine translation \cite{sutskever2014sequence, bahdanau2014neural}, speech recognition \cite{graves2013speech} and 
%large scale natrual language understanding \cite{kiros2015skip, hermann2015teaching}. 
%The success of RNNs has also
%inspired researchers to explore and better understand
%models with new model designs~\cite{chung2015gated, kalchbrenner2015grid,jozefowicz2015empirical}. 
Recently there has been a resurgence of new structural designs 
for recurrent neural networks (RNNs) \cite{chung2015gated,kalchbrenner2015grid,zhang2016architectural}.
Most of these designs are derived
from popular structures including vanilla RNNs, 
Long Short Term Memory networks (LSTMs)  \cite{hochreiter1997long} and Gated Recurrent Units (GRUs) \cite{cho2014learning}.
Despite of their varying characteristics, 
most of them share a common computational building
block, described by the following equation:
\begin{equation}
 \phi(\mathbf{W}\bm{x} + \mathbf{U}\bm{z}+ \mathbf{b}), 
\label{eq:rnn_basic}
\end{equation}
where $\bm{x} \in \mathbb{R}^n$ and $\bm{z} \in \mathbb{R}^m$ are state vectors
coming from different information sources,
$\mathbf{W} \in \mathbb{R}^{d\times n}$ and $\mathbf{U} \in \mathbb{R}^{d \times m}$
are state-to-state transition matrices, and $\bm{b}$ is a bias vector.
This computational building block
serves as a combinator for integrating information flow from the 
$\bm{x}$ and $\bm{z}$ by a sum operation
``$+$'', followed by a nonlinearity $\phi$.
We refer to it as the \textit{additive building block}.
Additive building blocks are widely implemented in various state computations in RNNs (e.g. 
%$\bm{h}_t = \phi(\mathbf{W}\bm{x}_t + \mathbf{U}\bm{h}_{t-1} + \mathbf{b})$ 
hidden state computations for vanilla-RNNs, gate/cell computations of LSTMs and GRUs.
%(in which we have 
%$\bm{g}_t = \phi([\mathbf{W}\bm{x}_t + \mathbf{U}\bm{h}_{t-1}] + \mathbf{b})$).

%\begin{center}
%\fbox{\parbox{0.3\textwidth}{
%\centering
%\vspace{5pt}
%$ \mathbf{W}\bm{x} + \mathbf{U}\bm{z}$ \\
%$\Downarrow$\\
% $ \mathbf{W}\bm{x}\odot \mathbf{U}\bm{z}$ \label{eq:3}
%\vspace{5pt}}}
%\end{center}

In this work, we propose an alternative design for constructing the computational building block
by changing the procedure of information integration.
Specifically, instead of utilizing sum operation ``+", we propose to use the Hadamard
product ``$\odot$'' to fuse $\mathbf{W}\bm{x}$ and $\mathbf{U}\bm{z}$:
\begin{equation}
 \phi(\mathbf{W}\bm{x} \odot \mathbf{U}\bm{z}+ \mathbf{b})
 \label{eq:sec_basic}
\end{equation}
The result of this modification changes the RNN from first order to second order \cite{goudreau1994first},
while introducing no extra parameters.
We call this kind of information integration
design a form of \textit{Multiplicative Integration}.
%A general formulation
%of Multiplicative Integration combines the
%first order integration by
%introducing extra biases and can be implemented in principle to integrate any information flows.
%In this section, we first propose the general formulation of the Multiplicative Integration,
%then we analysis its gradient flows and learning behaviors and show its relations
%to the previous works.
 %where $\mathbf{b} = \bm{\beta}_1\odot\bm{\beta}_2 + \bm{\beta}$.
%In Eq.~\ref{eq:sec_general}, $\bm{\alpha}$, $\bm{\beta}_1$
%and $\bm{\beta}_2$ can be considered as coefficients of 
%$\mathbf{W}\bm{x}_t \odot \mathbf{U}\bm{h}_{t-1}$, 
%$\mathbf{W}\bm{x}_t$ and $\mathbf{U}\bm{h}_{t-1}$
%that weight the contributions of each term inside $\phi$.
%If $\bm{\alpha} = \mathbf{0}$, Multiplicative Integration will degenerate to the original building block used in vanilla-RNNs, LSTMs and GRUs etc.. 
%If $\bm{\alpha}$ is learnable during training, and assume the model has enough learning #capabilities, then 
%the model can decide whether 
%it should degenerate to the regular RNN (i.e., $\bm{\alpha}=0$). This suggests that the training performance of the general will at least as same as the regular RNN.
The effect of multiplication
naturally results in a gating type structure, in which
$\mathbf{W}\bm{x}$ and $\mathbf{U}\bm{z}$
are the gates of each other.
%and the gate value is
%not restricted in range $[0, 1]$ due to the lack of nonlinearity. 
More specifically, one can think of the state-to-state computation 
$\mathbf{U}\bm{z}$ (where for example $\bm{z}$ represents the previous state)
as dynamically rescaled by $\mathbf{W}\bm{x}$ (where for example $\bm{x}$ represents the input).
Such rescaling  does not exist in the additive building block,
in which $\mathbf{U}\bm{z}$ is independent of $\bm{x}$.
This relatively simple modification brings about advantages
over the additive building block as it alters RNN's gradient properties, which 
we discuss in detail in the next section, as well
as verify through extensive experiments.

In the following sections, we first
introduce a general formulation of Multiplicative Integration. We then
compare it to the additive building block 
on several sequence learning tasks, including
character level language modelling, speech recognition,
 large scale sentence representation learning using a Skip-Thought model,
 and teaching a machine to read and comprehend for a question answering task.
 The experimental results (together with several existing state-of-the-art models) show 
that various RNN structures (including vanilla
RNNs, LSTMs, and GRUs)
equipped with Multiplicative Integration provide 
better generalization and easier optimization. Its main advantages include:
%\begin{itemize}
%\item 
(1) it enjoys better gradient properties due to the gating effect. Most of the hidden units are non-saturated;
(2) the general formulation of Multiplicative Integration naturally includes the regular additive building block as a special case,
and introduces almost no extra parameters compared to the additive building block;
and (3) it is a drop-in replacement for the additive building block
in most of the popular RNN models, including LSTMs and GRUs. It can also be combined with other RNN training techniques such as Recurrent Batch Normalization \cite{Cooijmans2016}. We further 
discuss its relationship to existing models, including Hidden Markov Models (HMMs) \cite{baum1967inequality}%, rabiner1989tutorial}
, second order RNNs \cite{ goudreau1994first} and Multiplicative RNNs \cite{sutskever2011generating}.% and the use of Multiplicative Integration in other areas.

\section{Structure Description and Analysis}
\label{sec:analysis}
%In this section, we first propose the general formulation of the Multiplicative Integration,
%then we analysis its gradient flows and learning behaviors and show its relations
%to the previous works.
\vspace{-5pt}
\subsection{General Formulation of Multiplicative Integration}
The key idea behind Multiplicative Integration is to integrate different 
information flows $\mathbf{W}\bm{x}$ and  $\mathbf{U}\bm{z}$, by the
Hadamard product ``$\odot$''.
A more general formulation of Multiplicative Integration includes two more bias 
vectors  $\bm{\beta}_1$ and $\bm{\beta}_2$ added to  
$\mathbf{W}\bm{x}$ and $\mathbf{U}\bm{z}$: 
\begin{equation}
 \phi((\mathbf{W}\bm{x}+\bm{\beta}_1)\odot(\mathbf{U}\bm{z} + \bm{\beta}_2) + \bm{b})
\label{eq:sec_gate} 
\end{equation}
where $\bm{\beta}_1, \bm{\beta}_2 \in \mathbb{R}^d$ are bias vectors.
Notice that such formulation contains the first order terms as in
a additive building block, i.e., $\bm{\beta}_1\odot\mathbf{U}\bm{h}_{t-1} + \bm{\beta}_2\odot  \mathbf{W}\bm{x}_t$.
%In order for the Multiplicative Integration to be flexible enough to have
%the additive building block as a special
%case, 
In order to make the Multiplicative Integration more flexible, 
we introduce another bias
vector $\bm{\alpha} \in \mathbb{R}^d$  
to gate\footnote{If $\bm{\alpha} = \mathbf{0}$, the Multiplicative Integration will
  degenerate to the vanilla additive building block.} the term $\mathbf{W}\bm{x} \odot \mathbf{U}\bm{z}$, obtaining 
the following formulation:
\begin{equation}
 \phi(\bm{\alpha}\odot \mathbf{W}\bm{x} \odot \mathbf{U}\bm{z}
								+ \bm{\beta}_1\odot \mathbf{U}\bm{z}
								+ \bm{\beta}_2\odot  \mathbf{W}\bm{x}+ \bm{b}),
\label{eq:sec_general}
\end{equation}
Note that the number of parameters of the Multiplicative Integration is about the same
as that of the additive building block, since the number of
new parameters ($\bm \alpha$, $\bm{\beta}_1$ and $\bm{\beta}_2$) are
negligible compared to total number of parameters. %(i.e., $O(md+nd)$).
Also, Multiplicative Integration can be easily extended to LSTMs and GRUs\footnote{See exact formulations in the Appendix.},
that adopt vanilla building blocks for computing gates and output states,
where one can directly replace them with the Multiplicative Integration.
More generally, in any kind of structure where $k$ information flows ($k \geq 2$) 
are involved (e.g. residual networks
\cite{he2015deep}), 
one can implement pairwise Multiplicative Integration for integrating all $k$ information sources.
%pairwisely for any subset of $k(k-1)/2$ information source pairs. 

\subsection{Gradient Properties}
\label{sec:grad_flow}
\vspace{-7.0pt}
The Multiplicative Integration has different gradient properties compared to the additive building block. 
For clarity of presentation, we first look at vanilla-RNN and 
RNN with Multiplicative Integration embedded, referred to as \textbf{MI-RNN}.
%In this case, the building block is embedded in the hidden state computation:
That is,
$\bm{h}_t = \phi(\mathbf{W}\bm{x}_t + \mathbf{U}\bm{h}_{t-1} + \mathbf{b})$ versus $\bm{h}_t = \phi(\mathbf{W}\bm{x}_t\odot\mathbf{U}\bm{h}_{t-1} + \mathbf{b})$.
In a vanilla-RNN, the gradient $\frac{\partial \bm{h}_t}{\partial \bm{h}_{t-n}}$
can be computed as follows:
\vspace{-5pt}
\begin{equation}
\label{eq:rnn_basic_grad}
\frac{\partial \bm{h}_t}{\partial \bm{h}_{t-n}} = \
\prod_{k = t-n+1}^{t}\mathbf{U}^{T}\mathrm{diag}(\phi'_k),
\label{eq:rnn_grad}
\end{equation}
where $\phi'_k = \phi'(\mathbf{W}\bm{x}_k + \mathbf{U}\bm{h}_{k-1} + \mathbf{b})$.
%In a additive building block, the properness of the gradient flow through time heavily depends
%on the hidden-to-hidden matrix $\mathbf{U}$, 
%while the input-to-hidden matrix $\mathbf{W}$ only plays limited roles
%since $\mathbf{W}\bm{x}_t$ are involved only as a bias term in hidden representation computation. 
The equation above shows that the gradient flow through time heavily depends
on the hidden-to-hidden matrix $\mathbf{U}$, but
$\mathbf{W}$ and $\bm{x}_k$ appear to play a limited role: 
they only come in the derivative of $\phi'$ mixed with $\mathbf{U}\bm{h}_{k-1}$. 
%Since
%$\mathbf{W}$ and $\bm{x}_t$ convey the information from input data, if they are not utilized well,
%then the model can hardly make meaningful gradient updates.
On the other hand, the gradient $\frac{\partial \bm{h}_t}{\partial \bm{h}_{t-n}}$
of a MI-RNN is\footnote{Here we adopt the simplest formulation of Multiplicative Integration for illustration. In the more general case (Eq.~\ref{eq:sec_general}),
$\mathrm{diag}(\mathbf{W}\bm{x}_k)$ in Eq.~\ref{eq:sec_raw_grad} will become $\mathrm{diag}(\bm{\alpha}\odot\mathbf{W}\bm{x}_k+\bm{\beta}_1)$.}:
\begin{equation}
\frac{\partial \bm{h}_t}{\partial \bm{h}_{t-n}} = \
\prod_{k = t-n+1}^{t}\mathbf{U}^{T}\mathrm{diag}(\mathbf{W}\bm{x}_k)\mathrm{diag}(\phi'_k),
\label{eq:sec_raw_grad}
\end{equation}
where $\phi'_k = 
\phi'(\mathbf{W}\bm{x}_k \odot \mathbf{U}\bm{h}_{k-1}+ \mathbf{b})$.
By looking at the gradient, we see that 
%the the result of the gating effect discussed earlier. 
the matrix $\mathbf{W}$ and the current input $\bm{x}_k$ is directly involved in the gradient computation by gating the matrix $\mathbf{U}$, hence more capable of altering the updates of the learning system. 
As we show in our experiments, with $\mathbf{W}\bm{x}_k$ directly gating the gradient, the vanishing/exploding problem is alleviated: $\mathbf{W}\bm{x}_k$ dynamically reconciles $\mathbf{U}$, making the gradient propagation easier compared to 
the regular RNNs. 
%By looking at the gradient, we clearly see the the result of the gating effect discussed earlier. The matrix $\mathbf{W}$ and the current input $\bm{x}_k$ is directly involved in the gradient computation by gating the matrix $\mathbf{U}$, hence more capable of altering the updates of the learning system. 
%In addition, with $\mathbf{W}\bm{x}_k$ directly gating the gradient, the vanishing/exploding problem is alleviated: $\mathbf{W}\bm{x}_k$ dynamically reconciles $\mathbf{U}$, making the gradient propagation more flexible and thus easier than merely multiplying the same $\mathbf{U}$ in
%vainlla-RNN. 
For LSTMs and GRUs with Multiplicative Integration,
the gradient propagation properties are more complicated.
But in principle, the benefits of the gating effect also persists in these models.

\vspace{-7.0pt}
\section{Experiments}
\vspace{-8pt}
In all of our experiments, we use the general 
form of Multiplicative Integration (Eq.~\ref{eq:sec_general}) for any hidden state/gate computations,
%and for LSTMs/GRUs we implement it to all the block input and gate computations,
unless otherwise specified.
\vspace{-8pt}
\subsection{Exploratory Experiments}
\vspace{-7.0pt}
To further understand the functionality of Multiplicative Integration,
%and explore the hypothesis proposed in Section~\ref{sec:analysis}, 
we take a simple RNN for illustration, 
and perform several exploratory experiments 
%which can be easily analysed without
%the interference of other complicated structural factors (like gates in LSTMs).
%The exploratory task is the
on the character level language modeling task using Penn-Treebank dataset \cite{marcus1993building},
following the data partition in \cite{mikolov2012subword}.
The length of the training sequence is 50. All models have
a single hidden layer of size~2048,
and we use Adam optimization algorithm~\cite{kingma2014adam} with learning rate $1e^{-4}$. Weights are initialized to samples drawn from
uniform$[-0.02, 0.02]$.
Performance is evaluated by the bits-per-character (BPC) metric,
which is $\log_2$ of perplexity.   
\vspace{-7.0pt}
\subsubsection{Gradient Properties}
\label{subsec:grad_pro}
\vspace{-7.0pt}
To analyze the gradient flow of the model, we divide the gradient in Eq.~\ref{eq:sec_raw_grad} into two parts: 
1. the gated matrix products:  $\mathbf{U}^{T}\mathrm{diag}(\mathbf{W}\bm{x}_k)$, 
and 2. the derivative of the nonlinearity $\phi'$, We separately analyze the properties of each term compared to the additive building block.
We first focus on the gating effect brought
by
$\mathrm{diag}(\mathbf{W}\bm{x}_k)$.
In order to separate out the effect of nonlinearity,
 we chose $\bm{\phi}$ to be the identity map, hence both vanilla-RNN and MI-RNN
reduce to linear models, 
referred to as \textbf{lin-RNN} and \textbf{lin-MI-RNN}. 

\begin{figure}[t]
\vspace{-10pt}
\centering
\minipage{0.4\textwidth}
  \includegraphics[width=\linewidth]{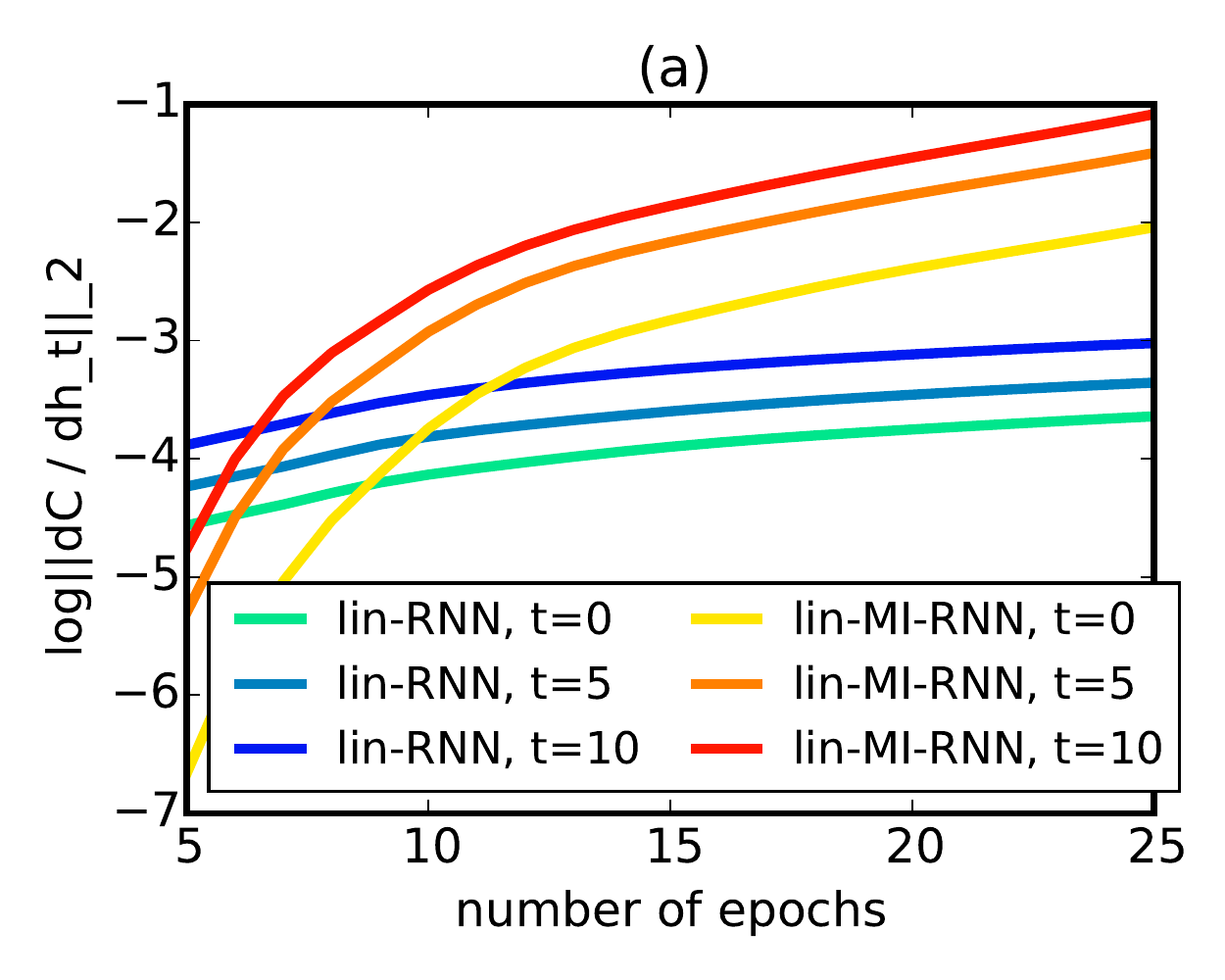}
\endminipage
\minipage{0.4\textwidth}%
  \includegraphics[width=\linewidth]{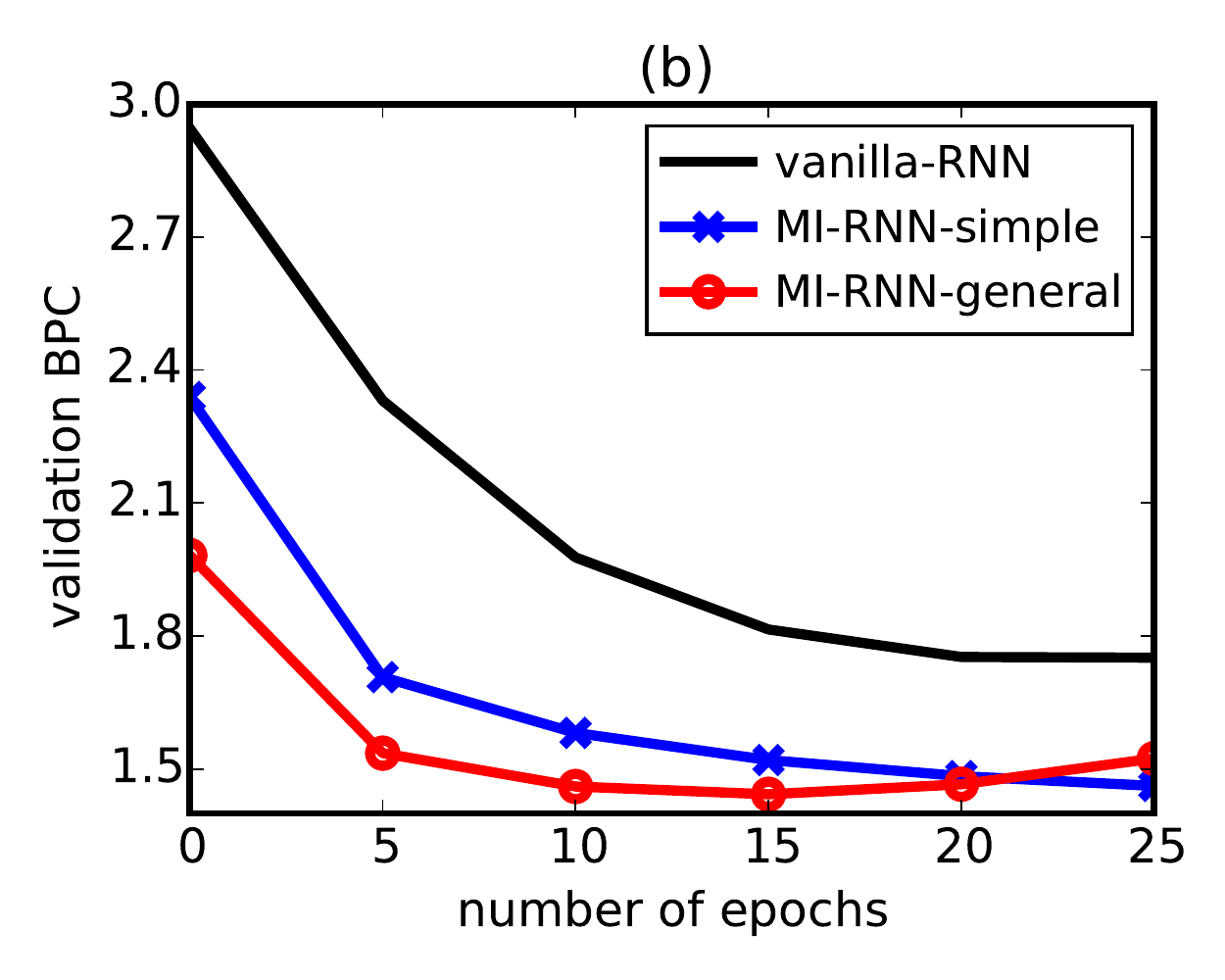}
\endminipage \\
\minipage{0.4\textwidth}
  \includegraphics[width=\linewidth]{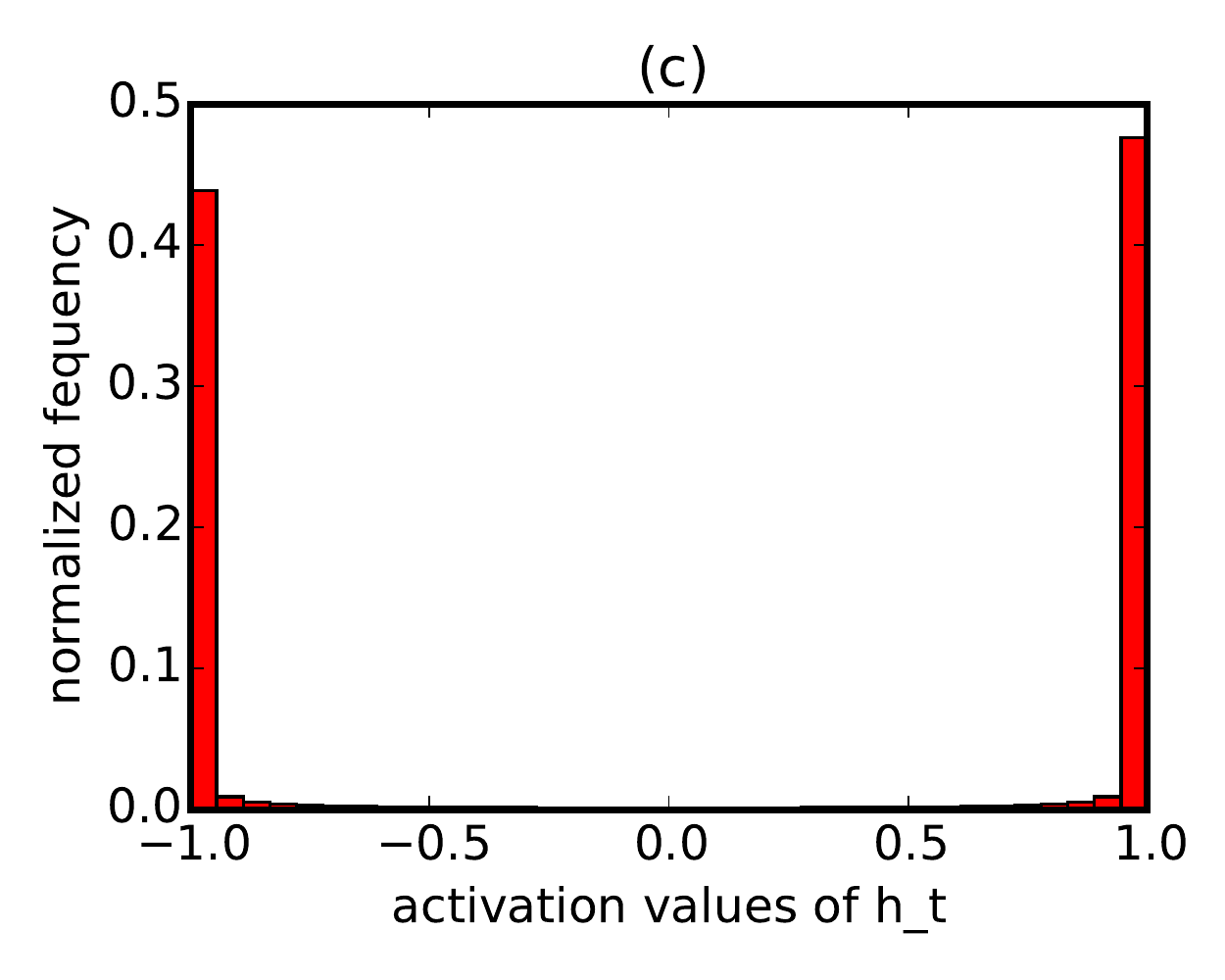}
\endminipage
\minipage{0.4\textwidth}
  \includegraphics[width=\linewidth]{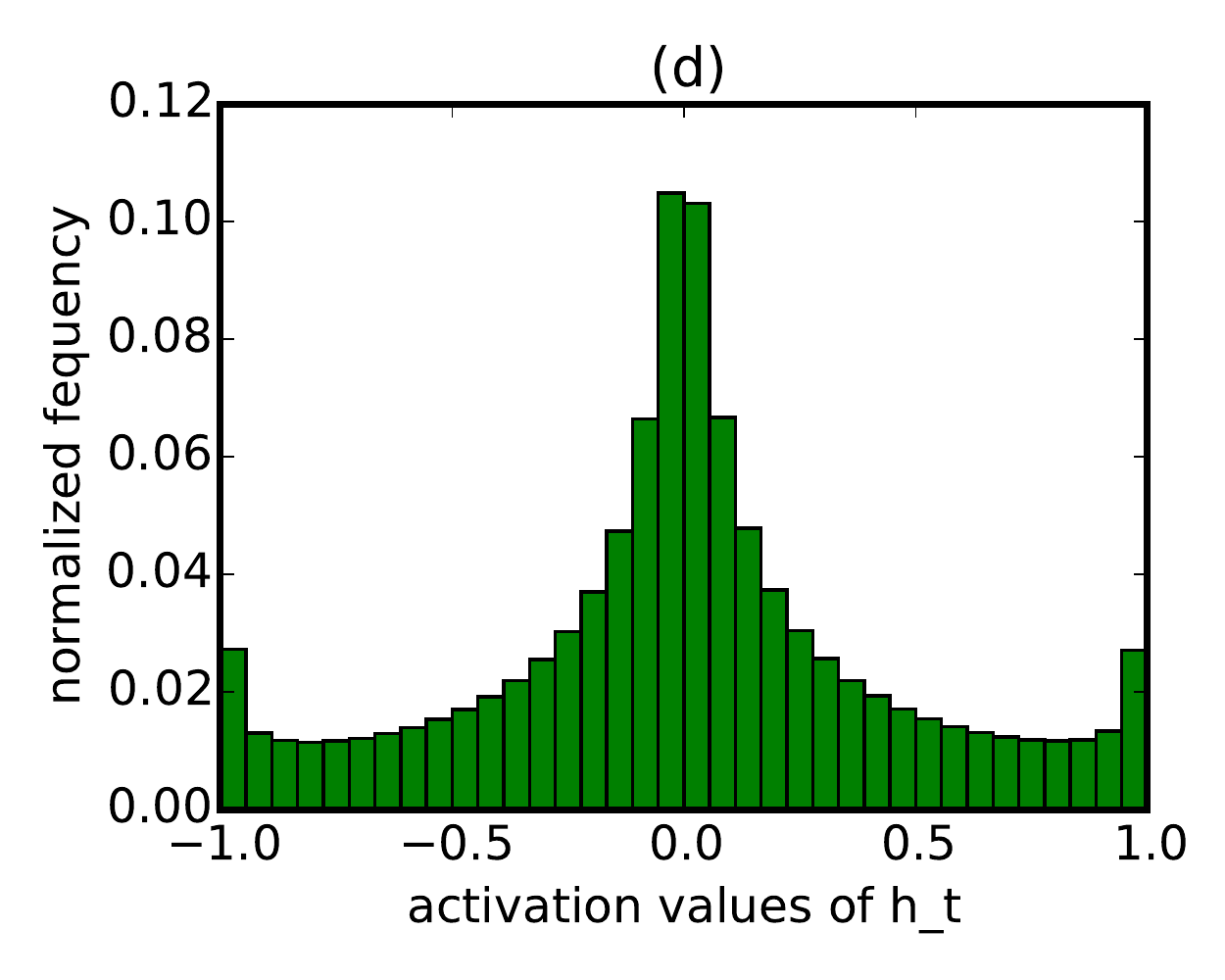}
\endminipage
\vspace{-10pt}
\caption{\small (a) Curves of log-L2-norm of
gradients for lin-RNN (blue) and lin-MI-RNN (orange).
Time gradually changes from \{1, 5, 10\}.
(b) Validation BPC curves for vanilla-RNN,
MI-RNN-simple using~Eq.~\ref{eq:sec_basic},
and MI-RNN-general using Eq.~\ref{eq:sec_general}.
%, MI-RNNs significantly
%outperform vanilla-RNN.
(c) Histogram of vanilla-RNN's hidden activations 
over the validation set, most activations are saturated.
(d) Histogram of MI-RNN's hidden activations 
over the validation set, most activations are \textbf{not} saturated.} 
\label{fig:explore}
\vspace{-10pt}
\end{figure}

For each model we monitor the log-L2-norm of the gradient 
$\log||\partial C/\partial \bm{h}_{t}||_2$ (averaged over the training set)
after every training epoch, where $\bm{h}_{t}$ is the hidden state at time step $t$, and $C$ is
the negative log-likelihood of the single character prediction at the final time step ($t = 50$). 
Figure.~\ref{fig:explore} shows the evolution of the gradient norms for small $t$, i.e., $0, 5, 10$, as they better
 reflect the gradient propagation behaviour.
Observe that the norms of lin-MI-RNN (orange)
increase rapidly and soon exceed the corresponding norms of 
lin-RNN by a large margin.
The norms of lin-RNN stay close to zero ($\approx10^{-4}$) and their changes over time are
almost negligible.
This observation implies that with the help of $\mathrm{diag}(\mathbf{W}\bm{x}_k)$ term,
the gradient vanishing of lin-MI-RNN can be alleviated compared to
 lin-RNN. The final test BPC (bits-per-character) of lin-MI-RNN is 1.48, which 
is comparable to a vanilla-RNN with stabilizing regularizer~\cite{krueger2015regularizing}, while lin-RNN performs rather poorly, achieving a test BPC of over~2.

Next we look into the nonlinearity $\phi$. We chose $\phi = \tanh$ for both vanilla-RNN and MI-RNN.
%referred to as \textbf{vanilla-RNN} and \textbf{MI-RNN}, 
%and monitor their hidden activations.
%Interestingly, most of the hidden activations in ``linMI-RNN'' are in
%non-saturated region whereas most activations in ``linRNN'' are saturated.
Figure~\ref{fig:explore} (c) and (d) shows a comparison of histograms of hidden
activations over all time steps on the validation set after training.
Interestingly, in (c) for vanilla-RNN, most activations are saturated with values around $\pm 1$,
whereas in (d) for MI-RNN, most activations are non-saturated with values around $0$.
This has a direct consequence in gradient propagation: non-saturated activations 
imply that $\mathrm{diag}(\phi'_k)\approx1$ for $\phi = tanh$, which 
can help gradients propagate, whereas saturated activations imply that 
$\mathrm{diag}(\phi'_k)\approx0$, resulting in gradients vanishing. 

%One possible explanation for such phenomenon is that multiplications of the hidden vectors are easier 

%``MI'' achieves test BPC of 1.39 which is
%state-of-the-art for simple RNN without complicated gating mechanisms, see Table~\ref{tb:bpc_penn} bottom left.
%To further verify that $diag(\bm{\alpha}\odot\mathbf{W}\bm{x}_k+\bm{\beta}_1)$ can better 
%control the gradient flows, 
%we run a linear version of these two structures which do not even contain element-wise nonlinearity $\phi$.
%Now the term $\phi_k$ does not exist from Eq.~\ref{eq:rnn_basic_grad} and Eq.~\ref{eq:sec_raw_grad}, 
%the linear regular RNN will surely explode/vanish
%due to the accumulated multiplications of the same $\mathbf{U}$, while situation can be better for Multiplication
%Integration because of the dynamic rescaling effect of $diag(\bm{\alpha}\odot\mathbf{W}\bm{x}_k+\bm{\beta}_1)$.
%The results demostrate our argument: the test BPC of regular linear RNN is very
%poor while the linear MI-RNN achieves surprisingly good result even like. Their gradient flow curves also is similar to .
\vspace{-5.0pt}
\subsubsection{Scaling Problem}
\vspace{-5.0pt}
When adding two numbers at different order of magnitude, the smaller one might be negligible for the sum. However, when
multiplying two numbers, the value of the product depends on both regardless of the scales.
This principle also applies when comparing Multiplicative Integration to the additive building blocks. %In regular RNNs, the quality of the training is very sensitive to the scales of $\mathbf{W}$ and $\mathbf{U}$: if these parameters are not carefully initialized, often after several epochs of training, one of the two terms in the sum becomes dominant, and the other becomes negligible.
In this experiment, we test whether Multiplicative Integration
is more robust to the scales of weight values. %, as discussed in Section~\ref{sec:scaling}.
Following the same models as in Section~\ref{subsec:grad_pro}, we first calculated the
norms of $\mathbf{W}\bm{x}_k$ and $\mathbf{U}\bm{h}_{k-1}$ for both vanilla-RNN and MI-RNN
for different $k$ after training. We found that in both structures,
$\mathbf{W}\bm{x}_k$ is a lot smaller than $\mathbf{U}\bm{h}_{k-1}$ in magnitude.
This might be due to the fact that $\bm{x}_k$ is a one-hot vector, making the number of updates
for (columns of) $\mathbf{W}$ be smaller than $\mathbf{U}$. As a result, in vanilla-RNN,
the pre-activation term $\mathbf{W}\bm{x}_k+\mathbf{U}\bm{h}_{k-1}$ is
largely controlled by the value of $\mathbf{U}\bm{h}_{k-1}$, while
$\mathbf{W}\bm{x}_k$ becomes rather small. 
% and play little roles for task prediction. 
In MI-RNN, on the other hand, the
pre-activation term $\mathbf{W}\bm{x}_k \odot \mathbf{U}\bm{h}_{k-1}$
still depends on the values of both $\mathbf{W}\bm{x}_k$ and $\mathbf{U}\bm{h}_{k-1}$, due to multiplication. 

%To explore this hypothesis even further, 
We next tried different initialization
of $\mathbf{W}$ and $\mathbf{U}$ to test their sensitivities to the scaling.
For each model, we fix the initialization of $\mathbf{U}$ to uniform$[-0.02, 0.02]$
and initialize $\mathbf{W}$ to uniform$[-r_{\mathbf{W}}, r_{\mathbf{W}}]$
where $r_{\mathbf{W}}$ varies in $\{0.02, 0.1, 0.3, 0.6\}$. 
Table~\ref{tb:bpc_penn}, top left panel, shows results. 
%The results shown in Table~\ref{tb:bpc_penn} Top exactly meets our expectation.
As we increase the scale of $\mathbf{W}$, 
performance of the vanilla-RNN improves, suggesting that the model is able to 
better utilize the input information. 
%we expect that the model can utilizes more information from the input, and the results show that this strategy helps for vanilla-RNN.
On the other hand, MI-RNN is much more robust to different initializations, where
the scaling has almost no effect on the final performance. 
%From the table we have
%that the performance variation of regular RNN with different initialization scales
%are large, while MI-RNNs' performances are quite stable as we expected.
%\begin{table}[!b]
%  \label{tb:bpc_penn}
%  \footnotesize
%  \centering
%  \begin{tabular}{c@{\hspace{4pt}c@{\hspace{4pt}}c@{\hspace{4pt}}c@{\hspace{4pt}}c@{\hspace{4pt}}c@{\hspace{4pt}}c}
%    \toprule
%    \{$r_{\mathbf{W}}$, $r_{\mathbf{U}}$\} & \{0.02, 0.02\} & \{0.02, 0.05\} &
%    \{0.02, 0.1\} & \{0.1, 0.02\} & \{0.3, 0.02\} & std\\
%    \hline
%    reg-RNN & 1.69 &   1.73 &  1.65 & 1.65 & 1.57 & 0.053\\
%    MI-RNN & 1.39 & 1.40 & - &1.40 & 1.40& \mathbf{0.004}\\
%    \bottomrule
%  \end{tabular}
%\end{table}
%\vspace{-10.0pt}
\subsubsection{On different choices of the formulation}
\vspace{-5.0pt}
In our third experiment, we evaluated the performance of 
different computational building blocks, which are
Eq.~\ref{eq:rnn_basic} (\textbf{vanilla-RNN}),
Eq.~\ref{eq:sec_basic} (\textbf{MI-RNN-simple})
and Eq.~\ref{eq:sec_general} (\textbf{MI-RNN-general})\footnote{We perform 
hyper-parameter search for the initialization of
$\{\bm{\alpha}, \bm{\beta}_1, \bm{\beta}_2, \mathbf{b}\}$ in MI-RNN-general.}. 
From the validation curves in Figure~\ref{fig:explore} (b), 
we see that both MI-RNN, simple and MI-RNN-general
yield much better performance compared to vanilla-RNN, and MI-RNN-general
has a faster convergence speed compared to MI-RNN-simple.
%, thanks to the extra freedom from
%$\bm{\alpha}$, $\bm{\beta}_1$ and $\bm{\beta}_2$. 
We also compared 
our results to the previously published models in
Table~\ref{tb:bpc_penn}, bottom left panel, 
where MI-RNN-general achieves a test BPC of 1.39, which is
to our knowledge the best result for RNNs on this task without complex gating/cell mechanisms.
%In the third experiment, we evaluate the influences of different initializations
%of $\bm{\alpha}$, $\bm{\beta}_1$ and $\bm{\beta}_2$ on MI model's performance.
%We compare regular RNN to 4 different MI models in which
%$\{\bm{\alpha}, \bm{\beta}_1, \bm{\beta}_2\}$
%are initialized as $\{0, 1, 1\}$, $\{1, 0, 0\}$, $\{1, 1, 1\}$ and $\{4, 0, 0\}$.
%From the validation curves in Fig.~\ref{} we observe that:
%1) MI-RNN with initialization $\{0, 1, 1\}$, 
%which at the beginning is structurally
%equivalent to regular RNN, 
%performs better than regular RNN. 
%2) The introduction of the second order term ($\alpha>0$) always
%helps for training which yields faster convergence and lower validation BPCs.
%3) MI-RNNs with larger $\alpha$s converge faster. 
%%We also observe the training acceleration effect of large $\alpha$ in 
%later experiments, though it is not always
%the case that larger $\alpha$ yields
%better validation/test performance.
%In fact, the best results from later experiments suggest that
%$\alpha$ should be within reasonable range of $[0, 2]$
%and for stacked RNNs ($\geq 2$ layers), $\alpha$ should be small
%compared to $\bm{\beta}_1, \bm{\beta}_2$. 
%and we want to check whether
%in Multiplicative Integration the gradient flow can be adjusted by $diag(\mathbf{W}\bm{x}_t+\mathbf{b}_2)\mathbf{U}$
%in a reasonable range, as we discussed in Section~\ref{sec:grad_flow}.
\begin{table}[!t]
  \footnotesize
  \centering
  \begin{tabular}{c@{\hspace{6pt}}c@{\hspace{4pt}}c@{\hspace{4pt}}c@{\hspace{4pt}}c@{\hspace{4pt}}c}
    \toprule
    $r_{\mathbf{W}}=$ & 0.02 & 0.1& 0.3 & 0.6 &std\\
    \midrule
    RNN & 1.69 & 1.65 & 1.57 & 1.54& 0.06\\
    MI-RNN & \textbf{1.39} & 1.40 & 1.40 & 1.41&0.008\\
    \bottomrule
  \end{tabular}
  \quad 
  \begin{tabular}{l@{\hspace{4pt}}c@{\hspace{4pt}}c}
    \toprule
    WSJ Corpus & CER & WER\\
    \midrule
    DRNN+CTCbeamsearch \cite{hannun2014first}& 10.0& 14.1 \\
    Encoder-Decoder \cite{bahdanau2015end}& 6.4& 9.3\\
    LSTM+CTCbeamsearch \cite{graves2014towards}& 9.2 & 8.7\\
    Eesen \cite{miao2015eesen}& - & \textbf{7.3}\\
    LSTM+CTC+WFST (ours)& 6.5 & 8.7\\
    MI-LSTM+CTC+WFST (ours)& \textbf{6.0} & 8.2\\
    \bottomrule
  \end{tabular}
  \\
  \vspace{7pt}
  \begin{tabular}{ll}
    \toprule
    Penn-Treebank     & BPC     \\
    \midrule
    RNN \cite{mikolov2012subword} & 1.42  \\
    HF-MRNN \cite{mikolov2012subword}   &  1.41\\
    RNN+stabalization \cite{krueger2015regularizing} & 1.48         \\
    MI-RNN (ours) &  \textbf{1.39}\\
    %\midrule
    %linear RNN (ours)  & >2\\
    linear MI-RNN (ours) & 1.48\\
    \bottomrule
  \end{tabular}
  \quad
  \begin{tabular}{ll}
    \toprule
    text8     & BPC     \\
    \midrule
    RNN+smoothReLu \cite{pachitariu2013regularization}  &1.55  \\
    HF-MRNN \cite{mikolov2012subword}    &  1.54\\
    MI-RNN (ours)   & \textbf{1.52}         \\
    \hdashline
    LSTM (ours) & $1.51$\\
    MI-LSTM(ours)& \textbf{1.44} \\
    %\midrule
    %LSTM-2000    & 1.43 \\
    %LSTM+BN-2000 &  \textbf{1.36}\\
    %MI-LSTM-2048 (ours) & 1.38\\
    \bottomrule
  \end{tabular}
  \quad
  \begin{tabular}{ll}
    \toprule
    HutterWikipedia     & BPC     \\
    \midrule
    stacked-LSTM \cite{graves2013generating} & 1.67  \\
    GF-LSTM \cite{chung2015gated}    &  1.58\\
    grid-LSTM \cite{kalchbrenner2015grid}  & 1.47         \\
    MI-LSTM (ours)& \textbf{1.44} \\
    \bottomrule
  \end{tabular}

\vspace{3pt}
\caption{\small Top: test BPCs and the standard deviation
of models with different scales of weight initializations.
Top right: test CERs and WERs on WSJ corpus.
Bottom left: test BPCs on character level Penn-Treebank dataset.
Bottom middle: test BPCs on character level text8  dataset.
Bottom right: test BPCs on character level Hutter Prize Wikipedia dataset.
}
\label{tb:bpc_penn}
\vspace{-20pt}
\end{table}

\subsection{Character Level Language Modeling}
\vspace{-5pt}
In addition to the Penn-Treebank dataset,
we also perform character level language modeling
on two larger datasets: \textit{text8}\footnote{\url{http://mattmahoney.net/dc/textdata}} and
\textit{Hutter Challenge Wikipedia}\footnote{\url{http://prize.hutter1.net/}}.
Both of them contain 100M characters from Wikipedia while \textit{text8} has
an alphabet size of 27 and  
\textit{Hutter Challenge Wikipedia} has an alphabet size of 205.
For both datasets, we follow the training protocols in~\cite{mikolov2012subword} 
and \cite{chung2015gated} respectively. 
We use Adam for optimization with the starting learning rate grid-searched
in $\{0.002, 0.001, 0.0005\}$. 
If the validation BPC (bits-per-character) does not decrease for 2 epochs, we half the
learning rate. %We initialize the weights from a uniform

We implemented Multiplicative Integration
on both vanilla-RNN and LSTM,
referred to as \textbf{MI-RNN} and \textbf{MI-LSTM}.
The results for the $text8$ dataset are shown in Table~\ref{tb:bpc_penn},
bottom middle panel. All five models, including some of the previously published models,  
have the same number of parameters ($\approx$4M).
For RNNs without complex gating/cell mechanisms (the first three results),
our MI-RNN (with $\{\bm{\alpha}, \bm{\beta_1},
\bm{\beta_2}, \mathbf{b}\}$ initialized as $\{2, 0.5, 0.5, 0\}$) performs the best, 
our MI-LSTM
(with $\{\bm{\alpha}, \bm{\beta_1}, \bm{\beta_2}, \mathbf{b}\}$ initialized as $\{1, 0.5, 0.5, 0\}$)
outperforms all other models by a large margin\footnote{\cite{Cooijmans2016} reports 
better results but they use much larger models ($\approx$16M)
which is not directly comparable.}.
% and achieves state-of-the-art
%of $\approx$4M model size

On \textit{Hutter Challenge Wikipedia} dataset, we compare our
MI-LSTM (single layer with 2048 unit, $\approx$17M,
with $\{\bm{\alpha}, \bm{\beta_1}, \bm{\beta_2}, \mathbf{b}\}$ initialized as $\{1, 1, 1, 0\}$) 
to the previous
stacked LSTM (7 layers, $\approx$27M) \cite{graves2013generating},
GF-LSTM (5 layers, $\approx$20M) \cite{chung2015gated},
and grid-LSTM (6 layers, $\approx$17M) \cite{kalchbrenner2015grid}.
Table~\ref{tb:bpc_penn}, bottom right
panel, shows results. Despite 
the simple structure compared to the sophisticated connection designs
in GF-LSTM and grid-LSTM, our MI-LSTM outperforms all other models and 
achieves the new state-of-the-art on this task. 

\subsection{Speech Recognition}
We next evaluate our models on Wall Street Journal (WSJ) corpus (available as LDC corpus LDC93S6B
and LDC94S13B), where we use the full 81 hour set ``si284'' for training,
set ``dev93'' for validation and set ``eval92'' for test.
%The raw audio is transformed
%into 40-dimensional log mel-filter-bank features with deltas and delta-deltas, which
%results in 120 dimensional features.
We follow the same data preparation process and model setting 
as in~\cite{miao2015eesen}, 
%the raw audio is transformed
%into 40-dimensional log mel-filter-bank features with deltas and delta-deltas, which
%results in 120 dimensional features,
and we use 59 characters as the targets for the acoustic modelling.
Decoding is done with the CTC~\cite{graves2006connectionist} based
weighted finite-state transducers (WFSTs)~\cite{mohri2002weighted} 
as proposed by~\cite{miao2015eesen}. 

Our model (referred to as \textbf{MI-LSTM+CTC+WFST})
consists of 4 bidirectional MI-LSTM layers, each with 320 units
for each direction. CTC is performed on top to resolve the alignment issue
in speech transcription. For comparison, we also train a baseline 
model (referred to as \textbf{LSTM+CTC+WFST}) with the same size but using vanilla LSTM.
Adam with learning rate $0.0001$ is used for optimization 
and Gaussian weight noise with zero mean and 
0.05 standard deviation is injected for regularization.
We evaluate our models on the character error rate (CER) without language model and the word
error rate (WER) with extended trigram language model.
%The comparsion between our model and the published results with similar building
%blocks is shown 

Table~\ref{tb:bpc_penn}, top right panel, shows that 
MI-LSTM+CTC+WFST achieves quite good results on both 
CER and WER compared to recent works,
and it has a clear improvement over the baseline model.
Note that we did not conduct a careful hyper-parameter search on this task, 
hence one could potentially obtain better results
with better decoding schemes and regularization techniques.

\subsection{Learning Skip-Thought Vectors}

Next, we 
evaluate our Multiplicative Integration on the Skip-Thought model of \cite{kiros2015skip}. 
Skip-Thought is an encoder-decoder model that attempts to learn 
generic, distributed sentence representations. 
The model produces sentence representation that are robust and perform well in practice, as it achieves excellent results across many different NLP tasks. 
The model was trained on the BookCorpus dataset that consists of 11,038 books with 
74,004,228 sentences. Not surprisingly, a single pass 
through the training data can take up to a week on a high-end GPU 
%(the required training time is very long, i.e., roughly two weeks 
(as reported in \cite{kiros2015skip}). Such training speed largely limits 
one to perform careful hyper-parameter search. 
However, with Multiplicative Integration, not only 
the training time is shortened by a factor of two, but the final performance 
is also significantly improved. %(**TODO** should we focus on speed here??) 

We exactly follow the authors' Theano implementation of the skip-thought 
model\footnote{\url{https://github.com/ryankiros/skip-thoughts}}: Encoder
and decoder are single-layer GRUs with hidden-layer size of 2400;
all recurrent matrices adopt orthogonal initialization while non-recurrent weights
are initialized from uniform distribution. Adam is used for optimization.
We implemented Multiplicative Integration
only for the encoder GRU (embedding MI into decoder did not provide any substantial gains).
We refer our model as \textbf{MI-uni-skip}, with $\{\bm{\alpha},\bm{\beta}_1,\bm{\beta}_2,\mathbf{b} \}$
initialized as $\{1,1,1,0\}$. 
We also train a baseline model with the same size, referred to as \textbf{uni-skip(ours)},
which essentially reproduces the original model of~\cite{kiros2015skip}.

\begin{figure}[t]
\vspace{-10pt}
\centering
\minipage{0.45\textwidth}
  \includegraphics[width=\linewidth]{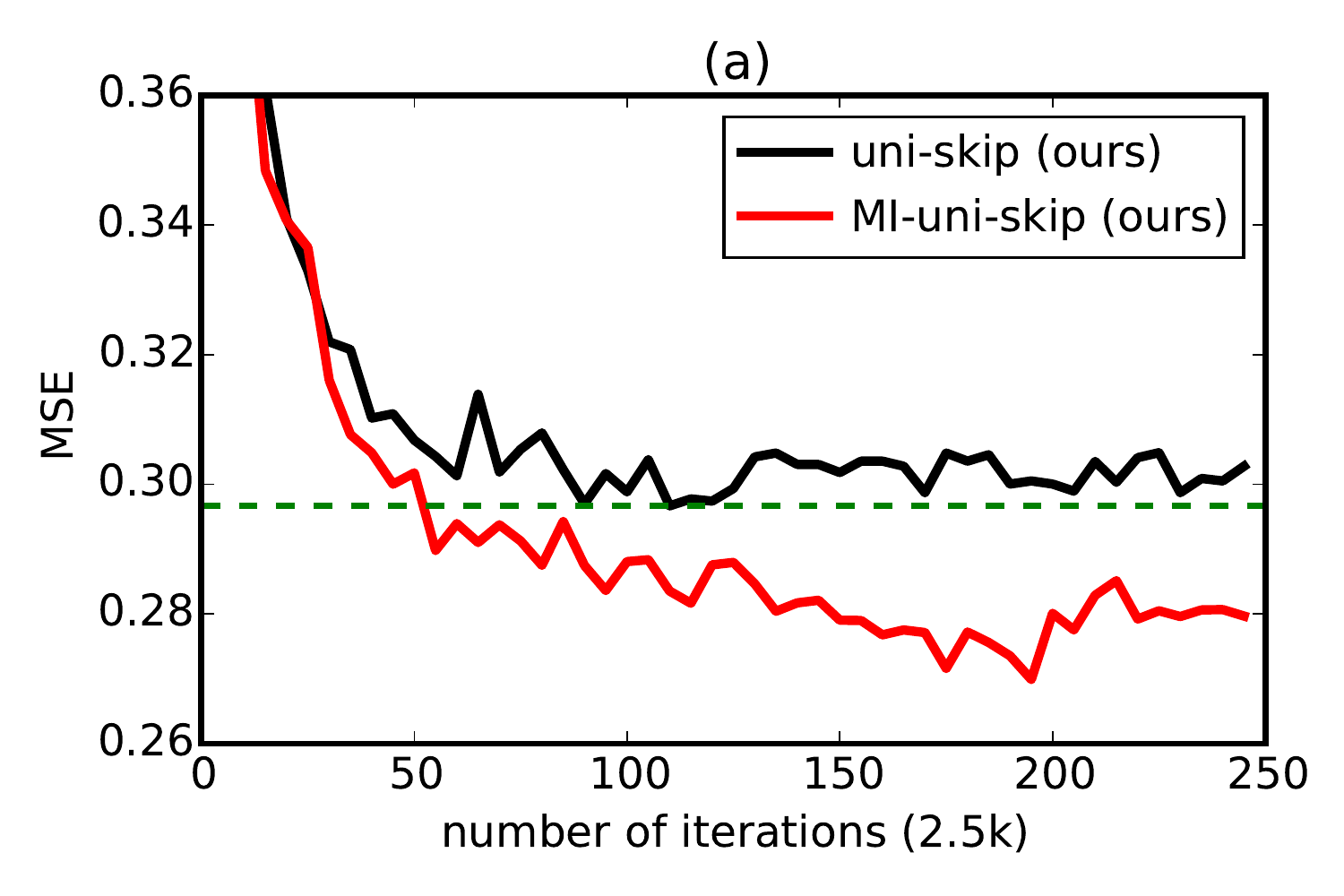}
\endminipage
\minipage{0.45\textwidth}%
  \includegraphics[width=\linewidth]{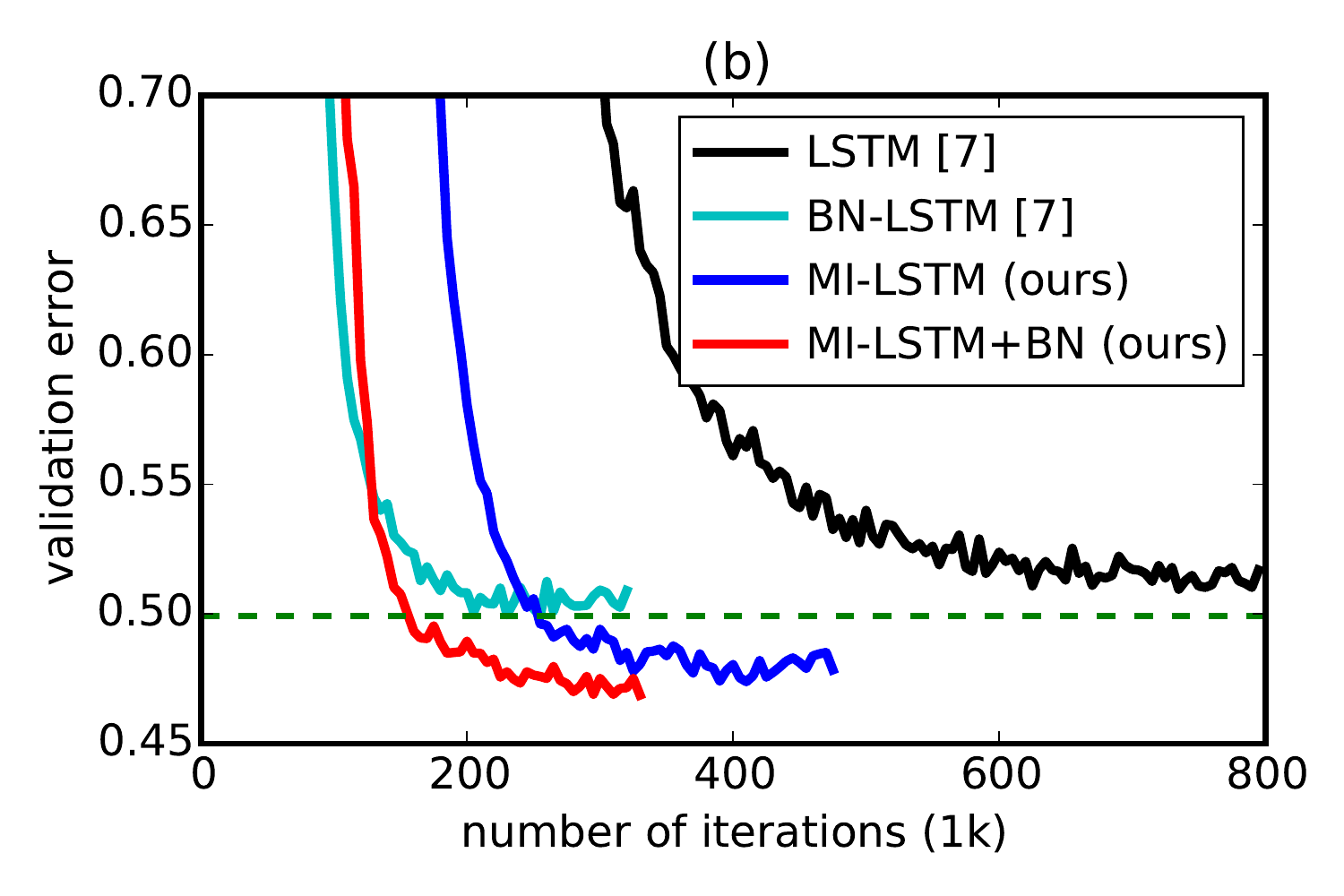}
\endminipage
\vspace{-10pt}
\caption{\small(a) MSE curves of uni-skip (ours) and MI-uni-skip (ours) on
semantic relatedness task on SICK dataset. MI-uni-skip significantly 
outperforms baseline uni-skip. 
(b) Validation error curves on attentive reader models. There is a clear margin 
between models with and without MI.} 
\label{fig:skip_reader}
\vspace{-10pt}
\end{figure}

During the course of training, we evaluated the skip-thought vectors 
on the semantic relatedness task, using SICK dataset, every 
2500 updates 
for both MI-uni-skip and the baseline model (each iteration processes a mini-batch of size 64). 
The results are shown in Figure~\ref{fig:skip_reader}a. 
Note that MI-uni-skip significantly outperforms 
the baseline, not only in terms of speed of convergence, but also 
in terms of final performance. 
At around 125k updates, 
MI-uni-skip already exceeds the best performance achieved by the baseline, 
which takes about twice the number of updates. 

We also evaluated both models after one week of training, with the best 
results being reported on six out of eight tasks reported in \cite{kiros2015skip}: 
semantic relatedness task on SICK dataset, 
paraphrase detection task on Microsoft Research Paraphrase Corpus, 
and four classification benchmarks: movie review sentiment (MR), 
customer product reviews (CR), subjectivity/objectivity
classification (SUBJ), and opinion polarity (MPQA). 
We also compared our results with the results reported 
on three models in the original skip-thought paper: 
\textbf{uni-skip}, \textbf{bi-skip}, \textbf{combine-skip}. Uni-skip 
is the same model as our baseline, bi-skip is a bidirectional 
model of the same size, and combine-skip takes the concatenation of
the vectors from uni-skip and bi-skip to form a 4800 dimension vector for task evaluation. 
Table \ref{tb:skipthought} shows that MI-uni-skip dominates across all the tasks. 
Not only it achieves higher performance than the baseline model, 
but in many cases, it also outperforms the combine-skip model, which has twice 
the number of dimensions. Clearly, Multiplicative Integration provides a faster and better way to 
train a large-scale Skip-Thought model. 
 
\begin{table}[t]
%\scriptsize
  \footnotesize
  %\fontsize{8pt}{12pt}
  \centering
\hspace{-15pt}
 \begin{tabular}{l@{\hspace{6pt}}c@{\hspace{4pt}}c@{\hspace{4pt}}c}
    \toprule
    Semantic-Relatedness  &  $r$ & $\rho$ & \textbf{MSE}       \\
    \midrule
    uni-skip \cite{kiros2015skip} &0.8477& 0.7780 &0.2872 \\
   bi-skip \cite{kiros2015skip} &0.8405 & 0.7696 &0.2995 \\
   combine-skip \cite{kiros2015skip} &0.8584& 0.7916 &0.2687 \\
    \midrule
    uni-skip (ours)  & 0.8436&  0.7735&  0.2946\\
    MI-uni-skip (ours) & \textbf{0.8588}&  \textbf{0.7952}&  \textbf{0.2679}\\
    \bottomrule
\end{tabular}
\quad 
\begin{tabular}{l@{\hspace{6pt}}c@{\hspace{4pt}}c}
    \toprule
    Paraphrase detection &  \textbf{Acc} & \textbf{F1}       \\
    \midrule
    uni-skip \cite{kiros2015skip} &73.0 & 81.9 \\
    bi-skip \cite{kiros2015skip} &71.2 & 81.2\\ 
    combine-skip \cite{kiros2015skip} &73.0 & 82.0 \\
    \midrule
    uni-skip (ours)  &\textbf{74.0}  &81.9  \\
    MI-uni-skip (ours) & \textbf{74.0}&  \textbf{82.1}\\
    \bottomrule
\end{tabular}
\\
\vspace{5pt}
\begin{tabular}{l@{\hspace{6pt}}c@{\hspace{4pt}}c@{\hspace{4pt}}c@{\hspace{4pt}}c}
    \toprule
    Classification &   MR & CR& SUBJ & MPQA       \\
    \midrule
    uni-skip \cite{kiros2015skip} &75.5& 79.3 &92.1& 86.9\\
    bi-skip \cite{kiros2015skip} &73.9& 77.9& 92.5& 83.3\\
    combine-skip \cite{kiros2015skip} &76.5& 80.1 &\textbf{93.6} &87.1\\
    \midrule
    uni-skip (ours)  & 75.9 &80.1 &  93.0 &87.0\\
    MI-uni-skip (ours) & \textbf{77.9} &  \textbf{82.3}&93.3&\textbf{88.1}\\
    \bottomrule
\end{tabular}
\quad
\begin{tabular}{ll}
    \toprule
    Attentive Reader     & Val. Err.     \\
    \midrule
    LSTM \cite{Cooijmans2016}  & 0.5033  \\
    BN-LSTM \cite{Cooijmans2016}    &  0.4951\\
    BN-everywhere \cite{Cooijmans2016}  & 0.5000         \\
    LSTM (ours)  & 0.5053  \\
    MI-LSTM (ours)     &  0.4721\\
    MI-LSTM+BN (ours)   & 0.4685         \\
    MI-LSTM+BN-everywhere (ours) &\textbf{0.4644} \\
   % \midrule
    %LSTM-512(ours) & 0.4644 \\
    %MI-LSTM-512(ours) & \textbf{0.4533} \\
    \bottomrule
  \end{tabular}
\vspace{3pt}
\caption{\small Top left: skip-thought+MI on Semantic-Relatedness task.
Top Right: skip-thought+MI on Paraphrase Detection task.
Bottom left: skip-thought+MI on four different classification tasks.
Bottom right: Multiplicative Integration (with batch normalization) on 
Teaching Machines to Read and Comprehend task.}
\label{tb:skipthought}
\vspace{-9pt}
\end{table}

\subsection{Teaching Machines to Read and Comprehend}
In our last experiment,
we show that the use of Multiplicative Integration can be combined 
with other techniques for training RNNs, and the advantages of using MI still persist.
Recently, \cite{Cooijmans2016} introduced Recurrent Batch-Normalization. They evaluated their proposed technique on a uni-directional 
Attentive Reader Model~\cite{hermann2015teaching} for the question answering task using 
the CNN corpus\footnote{Note that \cite{Cooijmans2016} used a truncated 
version of the original dataset in order to save computation.}.  
To test our approach, we evaluated the following four models: 
1. A vanilla LSTM attentive reader model with a single hidden layer size 240
(same as \cite{Cooijmans2016}) as our baseline, 
referred to as {\textbf{LSTM (ours)}}, 
2. A multiplicative integration LSTM with a single hidden size 240, 
referred to as {\textbf{ MI-LSTM}}, 
3. MI-LSTM with Batch-Norm, referred to as {\textbf {MI-LSTM+BN}}, 
4. MI-LSTM with Batch-Norm everywhere (as detailed in \cite{Cooijmans2016}), 
referred to as {\textbf {MI-LSTM+BN-everywhere}}. We compared our models 
to results reported in \cite{Cooijmans2016} (referred to as LSTM, BN-LSTM and BN-LSTM everywhere)
\footnote{Learning curves and the final result number are obtained by emails correspondence with authors of \cite{Cooijmans2016}.}. 

For all MI models,
$\{\bm{\alpha}, \bm{\beta}_1, \bm{\beta}_2, \mathbf{b}\}$ were initialized to $\{1,1,1,0\}$.
%We compared our models 
%with results by performing solely recurrent batch normalization\footnote{Learning curves and the final result number are obtained by emails correspondence with authors of \cite{Cooijmans2016}.}. 
We follow the experimental protocol of \cite{Cooijmans2016}\footnote{\url{https://github.com/cooijmanstim/recurrent-batch-normalization.git}.} 
and use exactly the same settings as theirs, except 
we remove the gradient clipping for MI-LSTMs. 
Figure.~\ref{fig:skip_reader}b shows validation curves of the baseline (LSTM), MI-LSTM, BN-LSTM, and MI-LSTM+BN, 
and the final validation errors of all models are reported in Table~\ref{tb:skipthought},
bottom right panel. 
Clearly, using Multiplicative Integration 
results in improved model performance regardless of whether Batch-Norm is used.  
%the best validation error is improved regardless of whether using Batch Norm. 
However, the combination of MI and Batch-Norm provides 
the best performance and the fastest speed of convergence. 
This shows the general applicability of Multiplication Integration 
when combining it with other optimization techniques. 

\section{Relationship to Previous Models}
 \vspace{-7.0pt}
\subsection{Relationship to Hidden Markov Models}
One can show that under certain constraints, MI-RNN
is effectively implementing the forward algorithm of the Hidden Markov Model(HMM). A direct mapping can be constructed as 
follows (see ~\cite{857908} for a similar derivation).
Let $\mathbf{U} \in \mathbb{R}^{m \times m}$ be the state transition probability matrix with $\mathbf{U}_{ij} = \mathrm{Pr}[h_{t+1} = i|h_{t} = j]$, $\mathbf{W} \in \mathbb{R}^{m\times n}$ be the
observation probability matrix with $\mathbf{W}_{ij} = \mathrm{Pr}[x_{t} = i|h_{t} = j]$. When $\bm{x}_t$ is a one-hot vector (e.g., in many of the language modelling tasks), multiplying it by $\mathbf{W}$ is effectively choosing a column of the observation matrix. Namely, if the $j^{th}$ entry of $\bm{x}_t$ is one, then $\mathbf{W}\bm{x}_t = \mathrm{Pr}[x_{t} |h_{t} = j]$.
Let $\bm{h}_0$ be the initial state distribution 
with $\bm{h}_0 = \mathrm{Pr}[h_{0}]$ and $\{\bm{h}_{t}\}_{t\geq1}$ be the alpha values in the forward algorithm of HMM, i.e., $\bm{h}_{t} = \mathrm{Pr}[x_{1},...,x_{t},h_{t}]$. Then $\mathbf{U}\bm{h}_{t} = \mathrm{Pr}[x_{1},...,x_{t},h_{t+1}]$. Thus $\bm{h}_{t+1} = \mathbf{W}\bm{x}_{t+1} \odot \mathbf{U}\bm{h}_{t} = \mathrm{Pr}[x_{t+1} |h_{t+1}]\cdot\mathrm{Pr}[x_{1},...,x_{t},h_{t+1}] = \mathrm{Pr}[x_{1},...,x_{t+1},h_{t+1}]$.
To exactly implement the forward algorithm using Multiplicative Integration, the matrices 
$\mathbf{W}$ and $\mathbf{U}$ have to be probability matrices, and $\bm{x}_t$ needs to be a one-hot vector. The function $\phi$ needs to be linear, and we drop all the bias terms. 
Therefore, RNN with Multiplicative Integration can be seen as a 
nonlinear extension of HMMs. The extra freedom in parameter values and nonlinearity makes the model more flexible compared to HMMs. %Similar idea can be found at~\cite{857908}.

\subsection{Relations to Second Order RNNs and Multiplicative RNNs}

MI-RNN is related to the second order RNN \cite{goudreau1994first} and the multiplicative RNN (MRNN)~\cite{sutskever2011generating}. 
We first describe the similarities with these two models:

The second order RNN involves a second order term $\bm{s}_t$ in a vanilla-RNN, where 
	the $i$th element $\bm{s}_{t,i}$ is computed by the bilinear form:
	%\begin{equation}
	$\bm{s}_{t,i} = \bm{x}_t^T\mathcal{T}^{(i)}\bm{h}_{t-1}$,
	%\label{eq:sec_sec} 
	%\end{equation}
	where $\mathcal{T}^{(i)}\in\mathbb{R}^{n\times m} (1\leq i\leq m)$ is
	the $i$th slice of a tensor $\mathcal{T}\in\mathbb{R}^{m\times n\times m}$.
	Multiplicative Integration also involve a second
	order term $\bm{s}_t = \bm{\alpha}\odot\mathbf{W}\bm{x}_t \odot \mathbf{U}\bm{h}_{t-1}$,
	but in our case 
	$\bm{s}_{t,i} = \alpha_i(\mathbf{w}_i\cdot\bm{x}_t)(\mathbf{u}_i\cdot\bm{h}_{t-1})
	= \bm{x}_t^T (\alpha\mathbf{w}_i\otimes\mathbf{u}_i) \bm{h}_{t-1}$,
where $\mathbf{w}_i$ and $\mathbf{u}_i$ are $i$th
row in $\mathbf{W}$ and $\mathbf{U}$, and $\alpha_i$ is the $i$th element
of $\bm{\alpha}$. Note that the outer product
$\alpha_i\mathbf{w}_i\otimes\mathbf{u}_i$ is a rank-1 matrix. % compared to the full-rank $\mathbf{U}_i$.
The Multiplicative RNN is also a second order RNN, but which approximates 
$\mathcal{T}$ by a tensor decomposition
$\sum \bm{x}_t^{(i)}\mathcal{T}^{(i)} = \mathbf{P}\mathrm{diag}(\mathbf{V}\bm{x}_t)\mathbf{Q}$.
%The matrix $\mathbf{U}$ is 	then dynamically depended on $\bm{x}_t$.
For MI-RNN, we can also think of the second order term
as a tensor decomposition: 
$\bm{\alpha}\odot \mathbf{W}\bm{x}_t \odot \mathbf{U}\bm{h}_{t-1}
=\mathbf{U}(\bm{x}_t)\bm{h}_{t-1} = [\mathrm{diag}(\bm{\alpha})\mathrm{diag}(\mathbf{W}\bm{x}_t)\mathbf{U}]\bm{h}_{t-1}$.
%and the hidden state computation becomes:
%\begin{equation}
%\bm{h}_t = \phi(\mathbf{W}\bm{x}_t + \mathbf{U}(\bm{x}_t)\bm{h}_{t-1} + \mathbf{b})  
%         = \phi(\mathbf{W}\bm{x}_t +  \mathbf{U}_{1}(\mathbf{W}^{mul}\bm{x}_t \odot \mathbf{U}_2\bm{h}_{t-1}) + \mathbf{b})
%\end{equation}

There are however several differences that make MI a favourable model:
%We summarize their differences as follows,
(1) Simpler Parametrization: 
 MI uses a rank-1 approximation compared to the second order RNNs,
and a diagonal approximation compared to Multiplicative RNN.
%MI-RNN uses rank-1 approximation
%for matrix $\mathbf{D}_i$ 
%compared to second order RNN, 
%replaces $\mathbf{P}$ with diagonal approximation $\mathrm{diag}(\bm{\alpha})$
%in the tensor decomposition compared to Multiplicative RNN.
Moreover, MI-RNN shares parameters across 
the first and second order terms, whereas the other two models do not.
As a result, the number of parameters are largely reduced,
which makes our model more practical for large scale problems, while 
avoiding overfitting.
%(2) Less parameters: 
%MI-RNN only has $O(nm+m^2)$ parameters which is almost the same as vanilla-RNN,
%while the second order RNN has $O(nm^2)$ parameters which is impractical for moderately large models, and MRNN also introduces
%at least two extra matrices compared to MI-RNN because of the tensor decomposition. 
%The obvious advantage of this is that MI can be practically implemented for large scale problems whereas the other two methods can't.  In the meantime, the model complexity is lowered which helps avoid overfitting.
(2) Easier Optimization: In tensor decomposition methods, 
the products of three different (low-rank) matrices generally makes it hard to optimize \cite{sutskever2011generating}. 
However, the optimization problem becomes easier in MI, as discussed in section 2 and 3.
%From the bilinear forms above we have
%that Multiplicative Integration is a special case of the
%second order RNN in which $\mathbf{D}$ shares parameters with $\mathbf{W}$ with $\mathbf{U}$
%by $\mathbf{D}_i= \alpha\mathbf{w}_i\otimes\mathbf{u}_i$.   
(3) General structural design vs. vanilla-RNN design: 
Multiplicative Integration can be easily 
embedded in many other RNN structures, e.g. LSTMs and GRUs,
whereas the second order RNN and MRNN present 
a very specific design for modifying vanilla-RNNs.

%General structural design versus vanilla-RNN design: In this paper we 
%focus on Multiplicative Integration's generality and emprically show that it can
%be embedded in many other RNN structures (e.g. LSTMs and GRUs) to achieve better performances, 
% whereas the second order RNN and MRNN are 
%specific modific on vanilla-RNNs.
%two different typfocuses on specific functionality design
%of $\mathbf{U}$ of the vanilla-RNN and they adopt quite different parameterization
%which introduces extra matrices because of the tensor factorization. Instead,
%Multiplicative Integration has simpler formulations and we focus on its structural
%generality for different RNN structures.
%Compared to Eq.~\ref{eq:sec_general}, if $\mathbf{U}_{1} = diag(\bm{\alpha})$ is diagonal
%and $\mathbf{W}^{mul} = \mathbf{W}$, the multiplicative RNN will degenerates to
%a special case of Multiplicative Integration with $\bm{\beta}_1 = \bm{0}$.
%Related ideas can be found in .
Moreover, we also compared MI-RNN's performance to the 
previous HF-MRNN's results (Multiplicative RNN trained by Hessian-free method)
in Table \ref{tb:bpc_penn}, bottom left and bottom middle panels, on Penn-Treebank and 
text8 datasets. 
One can see that MI-RNN outperforms HF-MRNN
on both tasks.
%For the other tasks, because of the huge computational cost, we drop MRNN or second order RNNs for comparisons.
\subsection{General Multiplicative Integration}
Multiplicative Integration can be viewed as a general way of 
combining
information flows from two different sources. 
In particular, \cite{journals/corr/RasmusVHBR15} proposed the ladder network that
achieves promising results on semi-supervised learning. In their model,
they combine the lateral connections and the backward connections 
via the ``combinator'' function by a Hadamard product.
The performance would severely degrade without this product as empirically shown by \cite{citeulike:13937173}.
\cite{yang-etal-2014} explored neural embedding approaches in knowledge bases
by formulating relations as bilinear and/or linear mapping
functions, and compared a variety of embedding models
on the link prediction task. Surprisingly, 
the best results among all bilinear functions 
is the simple weighted Hadamard product. They further carefully compare the multiplicative and additive interactions and show that the multiplicative interaction 
dominates the additive one. 
\vspace{-5pt}
\section{Conclusion}
\vspace{-2pt}
In this paper we proposed to use 
Multiplicative Integration (MI), a simple Hadamard product to 
combine information flow in recurrent neural networks.
MI can be easily integrated into many 
popular RNN models, including LSTMs and GRUs, while 
introducing almost no extra parameters.
Indeed, the implementation of MI requires almost no extra
work beyond implementing RNN models.
We also show that MI achieves state-of-the-art performance 
on four different tasks or 11 datasets of varying sizes and scales. 
We believe that the
Multiplicative Integration 
can become a default building block for training 
various types of RNN models.
%We also explore 
%the benefit MI brings about over additive building block in terms of gradient properties, and 
%verify them through extensive experiments. 
%\subsection*{Acknowledgments}
%Use unnumbered third level headings for the acknowledgments. All
%acknowledgments go at the end of the paper. Do not include
%acknowledgments in the anonymized submission, only in the final paper.
\section*{Acknowledgments}
The authors acknowledge the following agencies for funding and support: NSERC, Canada Research Chairs, CIFAR, Calcul Quebec, Compute Canada, Disney research and ONR Grant N00014-14-1-0232.
The authors thank the developers of Theano \cite{Bastien-Theano-2016} and Keras \cite{chollet2015}, and also thank Jimmy Ba for many thought-provoking discussions.

{\small 
\bibliography{nips_2016}
\bibliographystyle{unsrt}
}

\newpage
\appendix{\LARGE\textbf{Appendix}}
\section{Implementation Details}
\subsection{MI-LSTM}
Our MI-LSTM (without peephole connection) in experiments has the follow formulation:
\begin{eqnarray*}
&&\bm{z}_t = \mathrm{tanh}(\bm{\alpha}_z \odot \mathbf{W}_z\bm{x}_{t} \odot \mathbf{U}_z\bm{h}_{t-1} +
\bm{\beta}_{z,1} \odot \mathbf{U}_z\bm{h}_{t-1} + \bm{\beta}_{z,2} \odot \mathbf{W}_z\bm{x}_{t} + \mathbf{b}_z) \hspace{53pt}\textbf{block\ input} \\
&&\bm{i}_t = \sigma(\bm{\alpha}_i\odot \mathbf{W}_i\bm{x}_{t}\odot \mathbf{U}_i\bm{h}_{t-1} +
\bm{\beta}_{i,1}\odot\mathbf{U}_i\bm{h}_{t-1} + \bm{\beta}_{i,2}\odot\mathbf{W}_i\bm{x}_{t} + \mathbf{b}_i) \hspace{84pt}\textbf{input\ gate} \\
&&\bm{f}_t = \sigma(\bm{\alpha}_f\odot \mathbf{W}_f\bm{x}_{t}\odot \mathbf{U}_f\bm{h}_{t-1} +
\bm{\beta}_{f,1}\odot\mathbf{U}_f\bm{h}_{t-1} + \bm{\beta}_{f,2}\odot\mathbf{W}_f\bm{x}_{t} + \mathbf{b}_f) \hspace{66pt}\textbf{forget\ gate}\\
&&\bm{c}_t = \bm{i}_t\odot\bm{z}_t + \bm{f}_t\odot\bm{c}_{t-1}\hspace{296pt}\textbf{cell\ state}\\
&&\bm{o}_t = \sigma(\bm{\alpha}_o\odot \mathbf{W}_o\bm{x}_{t}\odot \mathbf{U}_o\bm{h}_{t-1} +
\bm{\beta}_{o,1}\odot\mathbf{U}_o\bm{h}_{t-1} + \bm{\beta}_{o,2}\odot\mathbf{W}_o\bm{x}_{t} + \mathbf{b}_o) \hspace{67pt}\textbf{output\ gate}\\
&&\bm{h}_t = \bm{o}_t\odot \mathrm{tanh}(\bm{c}_t)\hspace{301pt}\textbf{block\ output}
\end{eqnarray*}
where $\{\bm{\alpha}_*,\ \bm{\beta}_{*,1},\ \bm{\beta}_{*,2}\}_{* = z, i, f, o}$ are bias vectors, $\sigma$ denotes the sigmoid function.

%\section{miscellaneous}
\subsection{MI-GRU}
Our MI-GRU in experiments has the follow formulation:
\begin{eqnarray*}
&&\bm{z}_t = \sigma(\bm{\alpha}_z \odot \mathbf{W}_z\bm{x}_{t} \odot \mathbf{U}_z\bm{h}_{t-1} +
\bm{\beta}_{z,1} \odot \mathbf{U}_z\bm{h}_{t-1} + \bm{\beta}_{z,2} \odot \mathbf{W}_z\bm{x}_{t} + \mathbf{b}_z) \hspace{66pt}\textbf{update\ gate} \\
&&\bm{r}_t = \sigma(\bm{\alpha}_r\odot \mathbf{W}_r\bm{x}_{t}\odot \mathbf{U}_r\bm{h}_{t-1} +
\bm{\beta}_{r,1}\odot\mathbf{U}_r\bm{h}_{t-1} + \bm{\beta}_{r,2}\odot\mathbf{W}_r\bm{x}_{t} + \mathbf{b}_r) \hspace{78pt}\textbf{reset\ gate} \\
\nonumber&&\bm{\tilde{h}}_t = \mathrm{tanh}(\bm{\alpha}_h\odot \mathbf{W}_h\bm{x}_{t}\odot \mathbf{U}_h\left(\bm{r}_t \odot \bm{h}_{t-1}\right) +
\bm{\beta}_{h,1}\odot\mathbf{U}_h\left(\bm{r}_t \odot \bm{h}_{t-1}\right) + \bm{\beta}_{h,2}\odot\mathbf{W}_h\bm{x}_{t} + \mathbf{b}_h) \\
&&\hspace{351pt}\textbf{candidate activation}\\
&&\bm{h}_t = (1-\odot\bm{z}_t)\odot\bm{h}_{t-1} + \bm{z}_t\odot\bm{\tilde{h}}_{t-1}\hspace{207pt}\textbf{hidden activation}
\end{eqnarray*}
where $\{\bm{\alpha}_*,\ \bm{\beta}_{*,1},\ \bm{\beta}_{*,2}\}_{* = z, r, h}$ are bias vectors, $\sigma$ denotes the \rm{sigmoid} function.

%
%References follow the acknowledgments. Use unnumbered first-level
%heading for the references. Any choice of citation style is acceptable
%as long as you are consistent. It is permissible to reduce the font
%size to \verb+small+ (9 point) when listing the references. {\bf
%  Remember that you can use a ninth page as long as it contains
%  \emph{only} cited references.}
%\medskip
%
%\small
%
%[1] Alexander, J.A.\ \& Mozer, M.C.\ (1995) Template-based algorithms
%for connectionist rule extraction. In G.\ Tesauro, D.S.\ Touretzky and
%T.K.\ Leen (eds.), {\it Advances in Neural Information Processing
%  Systems 7}, pp.\ 609--616. Cambridge, MA: MIT Press.
%
%[2] Bower, J.M.\ \& Beeman, D.\ (1995) {\it The Book of GENESIS:
%  Exploring Realistic Neural Models with the GEneral NEural SImulation
%  System.}  New York: TELOS/Springer--Verlag.
%
%[3] Hasselmo, M.E., Schnell, E.\ \& Barkai, E.\ (1995) Dynamics of
%learning and recall at excitatory recurrent synapses and cholinergic
%modulation in rat hippocampal region CA3. {\it Journal of
%  Neuroscience} {\bf 15}(7):5249-5262.

\end{document}